\documentclass[3p]{elsarticle}

\newcommand{\xz}{\textcolor{red}}
\newcommand{\xzb}{\textcolor{blue}}
\newcommand{\xzg}{\textcolor{green}}

\pdfoutput=1
\usepackage{hyperref}
\usepackage{moreverb}
\usepackage{fancyhdr}
\usepackage{calc}
\usepackage{float}
\usepackage{subfigure}
\usepackage{color}
\usepackage[table]{xcolor}
\usepackage{amsmath}
\usepackage{arydshln}
\usepackage{ulem} 

\usepackage{graphicx}
\usepackage{epstopdf}

\usepackage{bm}
\usepackage{setspace}

\usepackage{enumerate}

\usepackage[ruled,vlined,linesnumbered,resetcount]{algorithm2e}

\usepackage{array}
\newcommand{\PreserveBackslash}[1]{\let\temp=\\#1\let\\=\temp}
\newcolumntype{C}[1]{>{\PreserveBackslash\centering}p{#1}}
\newcolumntype{R}[1]{>{\PreserveBackslash\raggedleft}p{#1}}
\newcolumntype{L}[1]{>{\PreserveBackslash\raggedright}p{#1}}

\usepackage{multirow}
\usepackage[figuresright]{rotating}

\journal{Journal of \LaTeX\ Templates}

\begin{document}

\begin{frontmatter}

\title{Benchmarking and Comparing Multi-exposure Image Fusion Algorithms}

\author[mymainaddress]{Xingchen Zhang\corref{mycorrespondingauthor}}

\cortext[mycorrespondingauthor]{Corresponding author}
\ead{xingchen.zhang@imperial.ac.uk}

\address[mymainaddress]{Department of Electrical and Electronic Engineering,
	Imperial College London\\ South Kensington Campus, Exhibition Road, London, SW7 2BT, UK}

\begin{abstract}
Multi-exposure image fusion (MEF) is an important area in computer vision and has attracted increasing interests in recent years. Apart from conventional algorithms, deep learning techniques have also been applied to multi-exposure image fusion. However, although much efforts have been made on developing MEF algorithms, the lack of benchmark makes it difficult to perform fair and comprehensive performance comparison among MEF algorithms, thus significantly hindering the development of this field. In this paper, we fill this gap by proposing a benchmark for multi-exposure image fusion (MEFB) which consists of a test set of 100 image pairs, a code library of 16 algorithms, 20 evaluation metrics, 1600 fused images and a software toolkit.~To the best of our knowledge, this is the first benchmark in the field of multi-exposure image fusion.~Extensive experiments have been conducted using MEFB for comprehensive performance evaluation and for identifying effective algorithms. We expect that MEFB will serve as an effective platform for researchers to compare performances and investigate MEF algorithms.
\end{abstract}

\begin{keyword}
multi-exposure image fusion, image fusion, benchmark, deep learning, image processing
\end{keyword}

\end{frontmatter}


\section{Introduction}
\begin{figure}
	\includegraphics[width=\textwidth]{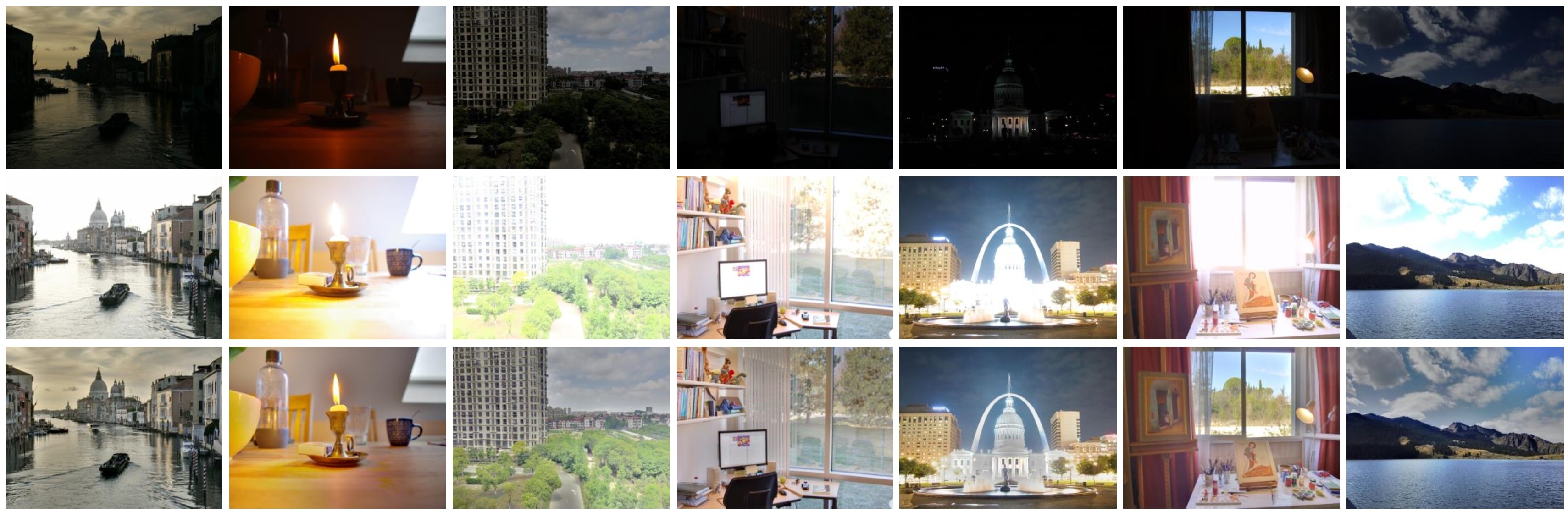}
	\caption{The benefit of multi-exposure image fusion.~The first row shows under-exposed images and the second row illustrates over-exposed images.~The third row gives the corresponding fused images produced by the MTI algorithm \cite{yang2018multi}.~As can be seen,the quality of images are greatly improved and more details are contained after fusion.}
	\label{fig:example-fusion}
\end{figure}

Due to the limited capture range of common imaging sensors, a single image captured by camera is usually insufficient to reveal all the details due to under-exposure or over-exposure.~The multi-exposure image fusion (MEF) technique can help to solve this problem by fusing information from multi-exposure images into an image \cite{shen2011generalized}, as shown in Fig.~\ref{fig:example-fusion}.

MEF has attracted wide attention due to its effectiveness in producing high-quality images and various MEF algorithms have been proposed.~Generally spearking, conventional MEF algorithms can be divided into spatial domain-based \cite{ma2015multi} and transform domain-based methods \cite{li2020fast}.~Spatial domain-based methods operate directly in spatial domain and can be roughly divided into two categories: pixel-based \cite{lee2018multi} and patch-based approaches \cite{ma2015multi}.~Besides, optimization-based MEF methods have also been developed \cite{ma2017multib}.~In contrast, transform domain-based methods firstly transform images into another domain and then perform fusion in that transformed domain.~The fused image is then obtained via inverse transformation.~However, these MEF approaches use hand-crafted features to fuse multi-exposure images, whose performances are limited since the hand-crafted features are not robust to varying input conditions.

In recent years, deep learning techniques have demonstrated great success in many computer vision areas, such as object tracking and detection, because of their very strong representation abilities.~Therefore, researchers have also tried to perform MEF using deep learning methods.~For instance, Lahoud et al.~\cite{lahoud2019fast} proposed a zero-learning image fusion algorithm based on pre-trained neural networks and demonstrated its effectiveness in MEF task.~Prabhakar et al.~\cite{prabhakar2017deepfuse} designed an unsupervised MEF algorithm named DeepFuse based on convolutional neural networks (CNN) using a no-reference quality metric as loss function.~Deng et al.~\cite{deng2020deep} presented a method based on a multi-modal convolutional sparse coding model which can be applied to several multi-modal image restoration and multi-modal image fusion problems.

\begin{table*}
	\begin{center}
		\caption{Some MEF algorithms published on top journals and conferences.~The number of tested image pairs, compared algorithms and utilized evaluation metrics are given.~The details of the proposed MEFB are also shown.~As can be seen, it is difficult to figure out which algorithm is really better since different test images, different metrics are chosen and compared with different MEF approaches.~Note that `$0$ metrics' means that there are only qualitative comparisons in the paper.}
		\label{table:published-some}
		\tiny
		\renewcommand\arraystretch{1.5}
		\begin{tabular}{llllll}		
			\hline
			Reference      & Year   & Journal/Conference     & Image pairs/sequences  & Algorithms  & Metrics \\ \hline
			\cite{shen2011generalized}  & 2011 & IEEE TIP  &  12         &      3       &  0  \\
			GFF \cite{li2013image}  &  2013   &   IEEE TIP                & 2 & 7  & 5 (MI, $Q_Y$, $Q_C$, $Q_G$, $Q_P$)\\
			DeepFuse \cite{prabhakar2017deepfuse} & 2017  & ICCV  & 23  &  7 & 1 (MEF-SSIM) \\ 
			SPD-MEF \cite{ma2017robust}   &  2017  & IEEE TIP   & 21  &  12  &  1 (MEF-SSIM) \\
			DIF-Net \cite{jung2020unsupervised}  & 2020 & IEEE TTP &24 & 6& 4 (MI, $Q_X$, $Q_H$, $Q_M$) \\
			IFCNN \cite{zhang2020ifcnn}  & 2020 & Information Fusion& 6 & 4   & 3 (MESSIM, SF, AG) \\
			FusionDN \cite{xu2020fusiondn}  & 2020  & AAAI & 30  & 5   &  4 (SD, EN, SSIM, VIF)       \\
			PMGI \cite{zhang2020PMGI}  & 2020 & AAAI & 19 & 3  & 6 (EN, SCD, CC, FMI, SD, MI) \\
			CU-Net \cite{deng2020deep}     & 2020   & IEEE TPAMI & 3   &  3   & 2 (PI, $Q_{NCIE}$)  \\		    
			FMMEF \cite{li2020fast}   & 2020  &IEEE TIP   & 21  & 9 & 1 (MEF-SSIM) \\      \hline
			\textbf{MEFB}                   & 2020 &      & \textbf{100} & \textbf{16} &\textbf{20} (CE, EN, FMI, MEF-SSIM, NMI, PSNR, $Q_{NCIE}$, TE,\\  
			&        &      &   &   &		   AG, EI, $Q^{AB/F}$, $Q_P$, SD, SF, $Q_C$, $Q_W$, $Q_Y$, $Q_{CB}$,  \\ 
			&        &      &   &   &		   $Q_{CV}$, VIF) \\ 
			\hline 
		\end{tabular}
	\end{center}
\end{table*}

However, although many MEF algorithms have been proposed, there is no benchmark available in this field which can be utilized to perform fair and comprehensive performance evaluation of MEF algorithms.~This is very different from other computer vision fields like object tracking where well-known benchmarks are widely used \cite{wu2013online, wu2015object}.~The lack of benchmark leads to several issues which significantly hinder the development of this field.~First, as shown in Table \ref{table:published-some}, normally different multi-exposure images are utilized in the literature as testing images.~Besides, different evaluation metrics and different compared algorithms are chosen for performance comparison.~These factors make it very difficult to conduct a really fair and comprehensive performance evaluation of MEF algorithms, thus hindering the understanding of the state-of-the-art of MEF field.~Second, in recent years, several deep learning-based MEF algorithms have been proposed which were claimed to have better performances.~However, since the absence of benchmark, there is a lack of extensive experiments which comprehensively compare the performance of deep learning-based algorithms and conventional MEF approaches.~Third, although some MEF algorithms are open-source, the interfaces and usage of these codes are different, making it time-consuming and non-trivial to conduct extensive experiments for performance comparison.~It is therefore desirable to have a platform which can be easily utilized to compare performances among algorithms.~It is also desirable to have fused images available which can be used directly.

To solve these issues, in this paper we present a multi-exposure image fusion benchmark (MEFB), which includes 100 pairs of multi-exposure images, 16 publicly available MEF algorithms, 20 evaluation metrics, 1600 fused images and a software toolkit to facilitate the algorithm running and performance evaluation.~The main contributions of this paper lie in the following aspects:

\begin{itemize}
	\item \textbf{Benchmark}.~A multi-exposure image fusion benchmark is proposed in this paper, including 100 pairs of multi-exposure images, 16 publicly available MEF algorithms, 20 evaluation metrics, 1600 fused images and a software toolkit.~To the best of our knowledge, this is the first benchmark in the field of multi-exposure image fusion.~The dataset in MEFB covers a wide range of environments thus is able to test the generalization ability of fusion algorithms.~20 evaluation metrics are implemented in MEFB to comprehensively compare fusion performance.~This is much more than those utilized in the MEF literature as shown in Table \ref{table:published-some}.~The provided toolkit can be easily used to run integrated algorithms and compute evaluation metrics.~It can also be easily extended to contain more test images, algorithms and evaluation metrics.
	
	\item \textbf{Comprehensive performance evaluation}.~Extensive experiments have been conducted using MEFB to compare the performances of MEF algorithms.~Specifically, deep learning-based MEF approaches are compared with conventional algorithms.
	~The dataset has been made publicly available \footnote{https://drive.google.com/file/d/1G2llZ8Te2Z8t0yKxGBOx7vtbpXcCVsNO/view?usp=sharing}.
\end{itemize}

The rest of this paper is organized as follows.~Section \ref{sec:related} gives some related work.~Then, the proposed multi-exposure image fusion benchmark is introduced in detail in Section \ref{sec:benchmark}, followed by experiments and results in Section \ref{sec:experiment}.~Section \ref{sec:discussions} gives some discussions.~Finally, Section \ref{sec:conclusion} concludes the paper.

\section{Related work}
\label{sec:related}
\subsection{Image Fusion}
Image fusion means combining information from multiple images into a single image which is more informative and better for downstream applications like tracking \cite{li2019rgb, zhang2020dsiammft, zhang2020object}.~Image fusion can be performed at pixel-level, feature-level and decision-level.~Also, it can either be performed in the spatial domain or transform domain.~According to the type of source images, there are several kinds of image fusion tasks, namely 
medical image fusion \cite{james2014medical, liu2017a}, multi-focus image fusion \cite{liu2017multi, zhang2020multifocus}, remote sensing image fusion \cite{ghassemian2016review, ye2018fusioncnn}, multi-exposure image fusion \cite{shen2011generalized, ma2015perceptual}, visible and infrared image fusion \cite{ma2016infrared, zhang2020vifb}.

Conventional image fusion methods mainly include weighted average method \cite{yin2018tensor}, wavelet transform based method \cite{hill2016perceptual}, PCA-based method \cite{he2010multimodal}, sparse representation method \cite{liu2015general} and compressed sensing method \cite{wan2011application}.~However, these algorithms are based on hand-crafted features.~In the past few years, a number of image fusion methods based on deep learning emerged \cite{jin2017survey, li2017pixel, liu2018deep, ma2019infrared} because of the strong representation abilities of deep learning techniques.~Regarding methods, CNN \cite{hermessi2018convolutional, liu2017multi, yan2018unsupervised, xia2018novel, prabhakar2017deepfuse}, generative adversarial networks (GAN) \cite{ma2019fusiongan}, Siamese networks \cite{liu2018infrared}, autoencoder \cite{li2019densefuse} have been explored to conduct image fusion.~It is foreseeable that more and more deep learning-based image fusion algorithms will emerge in the future.

\subsection{Performance evaluation of multi-exposure fusion algorithms}
It is non-trivial to evaluate MEF algorithms since the ground truth images are normally not available.~Generally speaking, there are two ways to evaluate MEF algorithms, namely subjective or qualitative method and objective or quantitative method  \cite{zhang2020ifcnn}.

Subjective evaluation means that the fused images are evaluated by human observers.~This is very useful in MEF research since a good fused image should be friendly to human visual system.~For example, Shen et al.~\cite{shen2011generalized} only used subjective evaluation to demonstrate the effectiveness of their method compared to others.~However, it is time-consuming and labor-intensive to observe each fused image in practice.~Besides, because each observer has different standard when observing fused images, thus biased evaluation may be easily produced.~Therefore, qualitative evaluation alone is not enough for the fusion performance evaluation.
~Quantitative method means using objective metrics to evaluate fusion algorithms.~Numerous evaluation metrics for image fusion have been proposed.~For instance, Ma et al.~\cite{ma2015perceptual} proposed a perceptual quality assessment metric (MEF-SSIM) for MEF method evaluation, Fang et al.~\cite{fang2019perceptual} developed an objective quality model for MEF of dynamic scenes.~However, none of these metrics is better than all other metrics.~As a consequence, in the literature different metrics are normally chosen as shown in Table \ref{table:published-some}, which makes the performance comparison under the same standard infeasible.

Apart from evaluation methods, test images are crucial to compare performances of MEF algorithms.~Some datasets have been proposed which can be used to test images \cite{ma2015perceptual, cai2018learning, liu2019perceptual}.~However, they do not provide code library and fusion results, thus it is inconvenient to perform large-scale experiments.~To the best of our knowledge, this is still not a benchmark in the field of multi-exposure image fusion, which provides dataset, code library, evaluation metrics, fused images and toolkit together.~The proposed MEFB aims to fill this gap.

\section{MEFB: A Multi-exposure image fusion benchmark}
\label{sec:benchmark}

\subsection{Dataset}
The MEFB dataset consists of 100 multi-exposure image pairs.~Each image pair contains an under-exposed image and an over-exposed image.~The dataset is a test set which is used for testing rather than training.~Note that in this study we focus on the fusion of one under-exposed image with an over-exposed one to obtain a photo-realistic natural image.~However, MEFB can be easily extended to include image sequences which contain more than two images.

\begin{figure}[H]
	\centering
	\includegraphics[width=\textwidth]{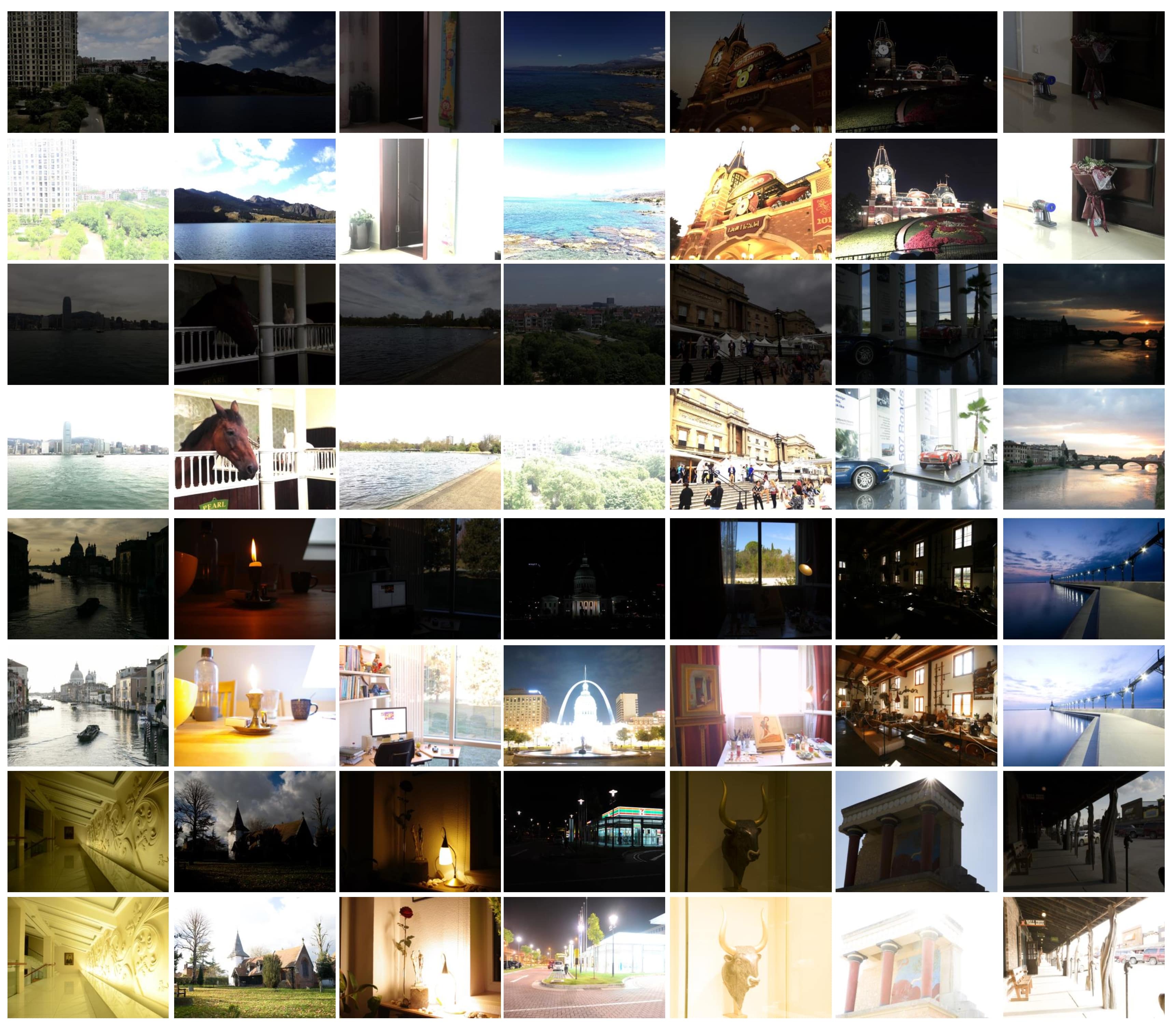}
	\caption{A part of test set in MEFB.~The first, third, fifth, and seventh rows show the under-exposed images while the second, fourth, sixth and eighth rows presents the corresponding over-exposed images.}
	\label{fig:dataset}
\end{figure}

The test set in MEFB comes from two sources.~The first part is collected from the Internet\footnote{https://ece.uwaterloo.ca/~k29ma/dataset/} and existing MEF datasets \cite{ma2015perceptual, zeng2014perceptual, liu2019perceptual, prabhakar2017deepfuse}.~The second part is captured by the authors of this study.~Since this paper aims to create a benchmark in the field of MEF, we think creating the test set in this way can maximize its value.~The reason is that in this way the images included in MEFB are captured with various cameras at various places, and thus covering a very wide range of  working environments and is thus suitable to test the robustness as well as generalization ability of MEF algorithms to varying inputs.~Besides, the images in MEFB have various resolutions so they can be used to test the ability of MEF algorithms to handle images with different resolutions.~This ability is very important for MEF approaches in real applications.~Figure \ref{fig:dataset} illustrate some examples of the dataset in MEFB.~As can be seen, many kinds of scenes such as indoor, outdoor, day and night are included in this dataset.~Besides, a lot kinds of objects are included like house, horse, candle, building, flower, river.

\subsection{Integrated algorithms}
At the moment, 16 algorithms are integrated in MEFB, including DEM \cite{wang2019detail}, DSIFT\_EF \cite{liu2015dense}, 
FMMEF \cite{li2020fast}, GD \cite{paul2016multi}, GFF \cite{li2013image}, IFCNN \cite{zhang2020ifcnn}, MEFAW \cite{lee2018multi}, MEFCNN \cite{li2018multib}, MEFDSIFT \cite{hayat2019ghost},  MEFNet \cite{ma2020deep}, MEFOpt \cite{ma2017multib}, MGFF \cite{bavirisetti2019multi}, 
MTI \cite{yang2018multi}, PMEF \cite{liu2019perceptual}, PWA \cite{ma2015multi}, SPD-MEF \cite{ma2017robust}.~In these algorithms, some were specifically designed for MEF, such as FMMEF and SPD-MEF, while some were originally designed for general image fusion including MEF, such as IFCNN and MGFF.~Besides, in these methods, IFCNN, MEFCNN and MEFNet are recent deep learning-based MEF methods published on top journals or conferences thus can represent the state-of-the-art deep learning-based MEF approaches.~More details about the categories of these algorithms can be founded in Table \ref{table:integrated}.

In the integrated algorithms, 13 of them are published in the last four years.~Besides, these 16 methods cover several kinds of MEF methods.~Therefore, the integrated methods can represent the state-of-the-art of MEF approaches.~Also, it should be mentioned that other MEF algorithms can be easily added to MEFB by either adding source codes using the designed toolkit interface or adding the fused images into the toolkit.

\begin{table*}
	\begin{center}
		\caption{Multi-exposure image fusion algorithms that have been integrated in MEFB.}
		\label{table:integrated}
		\footnotesize
		\renewcommand\arraystretch{1.5}
		\begin{tabular}{l|p{3cm}|p{8cm}}	
			\hline
			\multicolumn{2}{l|}{Category}         & Method  \\  \hline
			\multirow{2}{*}{Spatial domain-based}  & Pixel-based    &DSIFT\_EF \cite{liu2015dense}, MEFAW \cite{lee2018multi}, MEFDSIFT \cite{hayat2019ghost}, PMEF \cite{liu2019perceptual}\\ \cline{2-3}
			& Patch-based & PWA \cite{ma2015multi}, SPD-MEF \cite{ma2017robust}    \\ \cline{2-3}
			& Optimization-based     & MEFOpt \cite{ma2017multib} \\ \hline  
			\multirow{2}{*}{Transform domain-based} &  Multi-scale-based & DEM \cite{wang2019detail}, FMMEF \cite{li2020fast}, GD \cite{paul2016multi},   MTI \cite{yang2018multi}  \\ \cline{2-3}
			& edge-preserving-based& GFF \cite{li2013image}, MGFF \cite{bavirisetti2019multi}   \\ \hline                         			                                         
			\multicolumn{2}{l|}{Deep learning-based}     & IFCNN \cite{zhang2020ifcnn}, MEFNet \cite{ma2020deep}, MEMCNN \cite{li2018multi}  \\
			\hline
		\end{tabular}
	\end{center}
\end{table*}

\subsection{Evaluation metrics}
As introduced in \cite{liu2012objective}, image fusion evaluation metrics can be classified into four types as
\begin{itemize}
	\item Information theory-based
	
	\item Image feature-based
	
	\item Image structural similarity-based
	
	\item Human perception inspired
\end{itemize}

Zhang et al.~\cite{zhang2020vifb} demonstrated that image fusion algorithms may have significantly different performances on different kinds of metrics. ~Therefore, to have comprehensive performance comparison, 20 evaluation metrics were implemented in MEFB\footnote{The implementation of some metrics are kindly provided by Zheng Liu at https://github.com/zhengliu6699/imageFusionMetrics}.~Specifically, the implemented information theory-based metrics include 
cross entropy (CE) \cite{bulanon2009image}, entropy (EN) \cite{aardt2008assessment}, feature mutual information (FMI), normalized mutual information (NMI) \cite{hossny2008comments},  peak signal-to-noise ratio (PSNR) \cite{jagalingam2015review}, nonlinear correlation information entropy (Q$_{NCIE}$)  \cite{wang2005nonlinear, wang2008performance}, and tsallis entropy (TE) \cite{cvejic2006image}.~The implemented image feature-based metrics include average gradient (AG) \cite{cui2015detail}, edge intensity (EI) \cite{rajalingam2018hybrid}, gradient-based similarity measurement ($Q^{AB/F}$) \cite{xydeas2000objective}, phase congruency ($Q_P$) \cite{zhao2007performance}, standard division (SD)  \cite{rao1997fibre} and spatial frequency (SF) \cite{eskicioglu1995image}.~The implemented image structural similarity-based metrics include Cvejie's metric $Q_C$ \cite{cvejic2005similarity}, Peilla's metric ($Q_W$) \cite{piella2003new},  Yang's metric $(Q_Y)$ \cite{yang2008novel}, and MEF structural similarity index measure (MEF-SSIM) \cite{ma2015perceptual}.~The implemented human perception inspired fusion metrics are human visual perception ($Q_{CB}$) \cite{chen2009new}, $Q_{CV}$ \cite{chen2007human} and VIF \cite{han2013new}.~As can be seen, the evaluation metrics integrated in MEFB cover all four categories of metrics, thus are capable of quantitatively showing the quality of a fused image.~To the best of our knowledge, this is the first MEF work that implements as more as 20 metrics which cover all four kinds of evaluation metrics.

For all metrics except CE and  $Q_{CV}$, a larger value indicates a better fusion performance.~In MEFB, it is very convenient to compute all these metrics for each method, making it easy to compare performances of various algorithms.


Here we give the definitions of some evaluation metrics based on categories.~In the follow definitions, $M$ is the width of image, $N$ is the height of image.~$A$ and $B$ indicates the first (image A) and the second source image (image B), respectively.~$F$ represents the fused image.

\subsubsection{Information theory-based metrics}
\begin{enumerate}
	
	\item Cross entropy (CE).
	
	The CE between the fused image and all source images is defined as \cite{bulanon2009image}:
	\begin{equation}
	CE = \frac{CE_{A,F}+CE_{B,F}}{2},
	\end{equation}
	where $CE_{A,F}$ is the cross entropy between image A and the fused image, $CE_{B,F}$ is the cross entropy between image B and fused image.~$CE_{A,F}$ is computed as:
	\begin{equation}
	CE_{A,F} = \sum_{i=0}^{255}h_A(i)log_2\frac{h_A(i)}{h_F(i)},
	\end{equation}
	where $h(i)$ is the normalized histogram of the image.~$CE_{B,F}$ is computed as:
	\begin{equation}
	CE_{B,F} = \sum_{i=0}^{255}h_B(i)log_2\frac{h_B(i)}{h_F(i)}.
	\end{equation}
	A small CE value means the fused image has considerable similarity with source images, thus indicating a good fusion performance.
	
	\item Entropy (EN)  \cite{aardt2008assessment}.
	
	EN measures the information contained in the fused image \cite{aardt2008assessment}.~Its definition is:
	\begin{equation}
	EN = - \sum_{l=0}^{L-1}p_llog_2p_l,
	\end{equation}
	where $L$ represents the number of gray levels and $p_l$ denotes the normalized histogram of the corresponding gray level in the fused image.~A large EN indicates a bettor fusion performance.~However, EN is easily affected by noise, thus it is usually utilized as an auxiliary metric \cite{ma2019infrared}.
	
	\item Mutual information (MI) \cite{qu2002information}.\\
	MI is used to measure the amount of information that is transferred from source images to the fused image.~It is defined as:
	\begin{equation}
	MI = MI_{A, F} + MI_{B, F},
	\end{equation}
	where $MI_{A, F}$ and $MI_{B, F}$ denote the information transferred from image A and image B to the fused image, respectively.~Specifically, $MI_{X, F}$ is defined as follows:
	\begin{equation}
	MI_{X, F} = \sum_{x,f}p_{X, F}(x,f)log\frac{p_{X,F}(x,f)}{p_X(x)p_F(f)},
	\end{equation}
	where $X$ is $A$ for image A and is $B$ for image B, $p_X(x)$ and $p_F(f)$ are the marginal histograms of source image $X$ and fused image $F$, respectively.~$p_{X,F}(x,f)$ is the joint histogram of source image $X$ and fused image $F$.~A large MI value means a good fusion performance since considerable information is transferred to the fused image.
	
	\item Feature mutual information (FMI) \cite{haghighat2011non}.	
	
	FMI measures the amount of feature information that is transferred from source images based on MI and feature information.~It is defined as:
	\begin{equation}
	FMI = MI_{\hat{A}, \hat{F}} + MI_{\hat{B}, \hat{F}},
	\end{equation}
	where $\hat{A}, \hat{B}, \hat{F}$ are the feature maps of image A, image B and the fused image, respectively.
	
	A large FMI values indicates that more features are transferred from source images to the fused images, thus means a good fusion performance.
	
	\item Normalized mutual information (NMI) \cite{hossny2008comments}.
	
	NMI metric is defined as:
	\begin{equation}
	NMI =2\left[\frac{MI_{A, F}}{H(A)+H(F)} + \frac{MI_{B, F}}{H(B)+H(F)}\right]  ,
	\end{equation}
	where $H(\cdot)$ represents the entropy of the image.
	
	\item Peak signal-to-noise ratio (PSNR) \cite{jagalingam2015review}.
	
	PSNR indicates the ratio of peak value power and noise power in the fused image.~It can measure the distortion during the image fusion process and is defined as:
	\begin{equation}
	PSNR = 10log_{10}\frac{r^2}{MSE},
	\end{equation}
	where $r$ is the peak value of the fused image, $MSE$ is the mean squared error computed as:
	\begin{equation}
	MSE = \frac{MSE_{A, F}+MSE_{B,F}}{2},
	\end{equation}
	where $MSE_{A,F}=\frac{1}{MN}\sum_{i=0}^{M-1}\sum_{j=0}^{N-1}(A(i,j)-F(i,j))^2$, $MSE_{B,F}=\frac{1}{MN}\sum_{i=0}^{M-1}\sum_{j=0}^{N-1}(B(i,j)-F(i,j))^2$.~A large PSNR means that the fused image is close to source images and has less distortion.~Therefore, the larger the PSNR metric, the better the fusion performance is.	
	
	\item Nonlinear correlation information entropy (Q$_{NCIE}$)  \cite{wang2005nonlinear, wang2008performance}.
	
	$Q_{NCIE}$ is an information theory-based metric.~It constructs a nonlinear correlation matrix $R$ based on nonlinear correlation coefficient (NCC) between source images and the fused image, i.e.
	\begin{equation}
	R =  \left[
	\begin{matrix}
	1 & NCC_{A, B} & NCC_{A, F} \\
	NCC_{B, A} & 1 & NCC_{B, F} \\
	NCC_{F, A} & NCC_{F, B} & 1
	\end{matrix}
	\right].
	\end{equation}
	Then, $Q_{NCIE}$ can be computed as
	\begin{eqnarray}
	Q_{NCIE} = 1+\sum_{i=1}^{3} \frac{\lambda_i}{3}\log_{256}\frac{\lambda_i}{3},
	\end{eqnarray}
	where $\lambda_i$ are the eigenvalues of the matrix $R$.

	\item Tsallis entropy (TE) \cite{cvejic2006image}.
	TE is designed to measure the correlation between Tsallis entropy of source images and the fused image.~It is computed as
	\begin{equation}
	TE = I_{A, F}^\alpha (x, f) + I_{B, F}^\alpha(x, f),
	\end{equation}
	where $I_{X, F}$ exhibits the similarity of the Tsallis entropy between the source image $X$ and the fused image $F$.~It is defined as
	\begin{equation}
	I_{X, F}^\alpha = \frac{1}{\alpha - 1}(1-\sum_{x, f}\frac{(p_{X, F}(x, f))^\alpha}{(p_X(x)p_F(f))^{\alpha-1}}),
	\end{equation}
	where $p_{X, F}$ represents the joint distribution of the source image $X$ and the fused image $F$, $p_X(x)$ and $p_F(f)$ are the marginal histograms of source image $X$ and fused image $F$, respectively.  	
\end{enumerate}

\subsubsection{Image feature-based metrics}
\begin{enumerate}
	
	\item Average gradient (AG) \cite{cui2015detail}.
	
	AG measures the gradient information of the fused image and represents its detail and texture.~It is defined as:
	\begin{equation}
	AG = \frac{1}{MN}\sum_{i=1}^{M}\sum_{j=1}^{N}\sqrt{\frac{\nabla F_x^2(i,j)+\nabla F_y^2(i,j)}{2}},
	\end{equation}
	where $\nabla F_x(i,j) = F(i,j)-F(i+1,j)$, $\nabla F_y(i,j) = F(i,j)-F(i,j+1)$.~A large AG value indicates that more gradient information is contained in the fused image and thus means a good fusion performance.

	\item Edge intensity (EI) \cite{rajalingam2018hybrid}.
	
	EI measures the edge intensity information of an image.~A higher EI value indicates more clearness and higher image quality.~EI can be computed using Sobel operator as:
	\begin{equation}
	EI = \sqrt{S_x^2 + S_y^2},	
	\end{equation}
	where 
	\begin{equation}
	S_x = F*h_x, S_y = F*h_y,
	\end{equation}
	where $h_x = 
	\left[
	\begin{matrix}
	1 & 0 & 1 \\
	-2 & 0 & 2 \\
	-1 & 0 & 1
	\end{matrix}
	\right]$,  $h_y = 
	\left[
	\begin{matrix}
	1 & -2 & -1 \\
	0 & 0 & 0 \\
	1 & 2 & 1
	\end{matrix}
	\right]$, $*$ is convolution operation.
	
	\item Edge based similarity measurement ($Q^{AB/F}$) \cite{xydeas2000objective}.
	
	$Q^{AB/F}$ indicates the amount of edge information that is transferred form source images to fused image.~It can be computed as:
	\begin{equation}
	Q^{AB/F} = \frac{\sum_{i=1}^{N}\sum_{j=1}^{M}(Q^{A,F}(i,j)w^A(i,j)+Q^{A,F}(i,j)w^B(i,j))}{\sum_{i=1}^{N}\sum_{j=1}^{M}(w^A(i,j)+w^B(i,j))},
	\end{equation}
	where $Q^{X,F}(i,j) = Q_g^{X,F}(i,j)Q_a^{X,F}(i,j)$, $Q_g^{X,F}(i,j)$ and $Q_a^{X,F}(i,j)$ represents the edge strength and orientation values at location $(i,j)$, respectively.~$w^X$ denotes the weight that expresses the importance of each source image to the fused image.~Here, $X$ is $A$ for image A and is $B$ for image B.~The more edge information transferred to the fused image, the larger the $Q^{AB/F}$ value is.~Thus a large $Q^{AB/F}$ value indicates a good fusion performance.
	
	\item Standard devision (SD)  \cite{rao1997fibre}. 
	
	SD reflects the distribution and contrast of the fused image.~Its definition is:
	\begin{equation}
	SD= \sqrt{\sum_{i=1}^{M}\sum_{j=1}^{N}(F(i,j)-\mu)^2}
	\end{equation}
	where $\mu$ represents the mean value of the fused image.~The human visual system is sensitive to contrast, thus the regions in an image with high contrast always attract human attention.~Since high contrast in a fused image leads to a large SD, thus a large SD indicates that the fused image is with a good visual effect.

	\item Spatial frequency (SF) \cite{eskicioglu1995image}.\\
	SF \cite{eskicioglu1995image} can measure the gradient distribution of an image thus revealing the detail and texture of an image.~It is defined as:
	\begin{equation}
	SF = \sqrt{RF^2+CF^2},
	\end{equation} 
	where $RF = \sqrt{\sum_{i=1}^{M}\sum_{j=1}^{N}(F(i,j)-F(i,j-1))^2} $ and $CF = \sqrt{\sum_{i=1}^{M}\sum_{j=1}^{N}(F(i,j)-F(i-1,j))^2}$.~A large SF value indicates rich edges and textures, thus indicating good fusion performance.

\end{enumerate}

\subsubsection{Image structural similarity-based metrics}
\begin{enumerate}

	\item Yang's metric $(Q_Y)$ \cite{yang2008novel}.
	
	$Q_Y$ is a SSIM \cite{wang2004image}-based fusion quality metric.~It indicates the amount of structural information retained in the fused image $F$ from both source images.~$Q_Y$ is defined as:
	\begin{equation}
	Q_Y =\left\{ 
	\begin{aligned}
	&\lambda(w) SSIM(A,F|w) + (1-\lambda(w))SSIM(B,F|w),~~~ if~~ SSIM(A,B|w) \geq 0.75 \\
	&max(SSIM(A,F|w), SSIM(B,F|w)), ~~~~~~~~~~~~~~~~~if~~ SSIM(A,B|w) < 0.75
	\end{aligned}
	\right.
	\end{equation}
	where $w$ is a local window and $\lambda(w)$ is
	\begin{equation}
	\lambda(w) = \frac{s(A|w)}{s(A|w)+s(B|w)},
	\end{equation}
	where $s$ is a local measure of image saliency.
	
	\item Structural similarity index measure (SSIM) \cite{wang2004image}.
	
	SSIM is used to model image loss and distortion, to which the human visual system is sensitive.~It consists of three parts, namely loss of correlation, luminance, and contrast distortion.~SSIM between a source image and the fused image is defined as the product of these three parts, i.e.
	\begin{equation}
	SSIM_{X,F} = \sum_{x,f}\frac{2\mu_x\mu_f+C_1}{\mu_x^2+\mu_f^2+C_1}\cdot\frac{2\mu_x\mu_f+C_2}{\mu_x^2+\mu_f^2+C_2}\cdot\frac{\sigma_{xf}+C_3}{\sigma_x\sigma_f+C_3},
	\end{equation}
	where $SSIM_{X,F}$ denotes the structural similarity between source image $X$ ($X$ is $A$ for image A and is $B$ for image B) and fused image $F$, $x$ and $f$ represent the image patches of source and fused image in a sliding window, respectively.~$\sigma_{xf}$ is the covariance of source and fused images, $\sigma_x$ and $\sigma_f$ represent the standard deviation, $\mu_x$ and $\mu_f$ are the mean values of source and fused images, respectively.~$C_1, C_2, C_3$ are the parameters used to make the algorithm stable.
	
	The structural similarities between the fused image and both source images can be defined as:
	\begin{equation}
	SSIM = SSIM_{A,F} +SSIM_{B,F}.
	\end{equation}
	
	A large SSIM value indicates a better fusion performance.
	
\end{enumerate}

\subsubsection{Human perception inspired fusion metrics}
\begin{enumerate}
	
	\item  Human visual perception ($Q_{CB}$ ) \cite{chen2009new}.
	
	$Q_{CB}$ mainly measures the similarity of the major features in the human visual system.~It is computed as
	\begin{equation}
	Q_{CB} = \frac{1}{MN} \left( \sum_{i=1}^{N} \sum_{j=1}^{M}\lambda_A(i,j) Q_{A, F}(i,j) + \lambda_B(i,j) Q_{B, F}(i,j)\right) ,
	\end{equation}
	where $Q_{A, F}(i,j)$ and $Q_{B, F}(i,j)$ indicates the contrast transformed from source images to the fused image, $\lambda_A$ and $\lambda_B$ are the saliency maps of  $Q_{A, F}(i,j)$ and $Q_{B, F}(i,j)$, respectively.

	\item Visual information fidelity (VIF) \cite{han2013new}
	
	VIF measures the information fidelity of the fused image, which is consistent with the human visual system.~It aims to build a model to compute the distortion between the source and fused images.~The computation of VIF consists of four steps: First, the source images and fused image are filtered and divided into different blocks.~Second, the visual information of each block with and without distortion is evaluated.~Third, the VIF for each subband is calculated. Finally, the overall metric based on VIF is calculated.~A larger VIF value means a better fusion performance. 
	
\end{enumerate}

More information about evaluation metrics can be founded in \cite{liu2012objective, jagalingam2015review,  ma2015perceptual, ma2019infrared}.

\subsection{The software toolkit}
The aim of the toolkit in MEFB is to facilitate algorithm running and performance evaluation.~We have designed three interfaces in the toolkit.~The first one is used to add test images and fused images.~The second one is used to add MEF algorithms.~The third one is used to add new metrics. By using these interfaces, it is easy to extend MEFB.~Also, with this toolkit, running MEF algorithms and computing metrics are as easily as pushing a button once settings are done.

Users can use this toolkit in two ways.~First, they can add their algorithms into MEFB, and then produce the fused
images and compute metrics using the toolkit.~Second, they can produce fused images outside this toolkit if they want and put
their fused images into MEFB to compute metrics.~Then, their algorithms can be compared with those methods in MEFB.

\section{Experiments and results}
\label{sec:experiment}
This section presents experimental results obtained with MEFB.~All experiments were performed using a computer equipped with an NVIDIA RTX2070 GPU and i7-9750H CPU.~All fused images were generated by the publicly available implementations with default settings.~Note that pre-trained models of each deep learning algorithm were provided by the corresponding authors of each algorithm.~The dataset in MEFB was only used for performance evaluation of those algorithms but not for the training.

\begin{figure}[H]
	\centering
	\includegraphics[width=\textwidth]{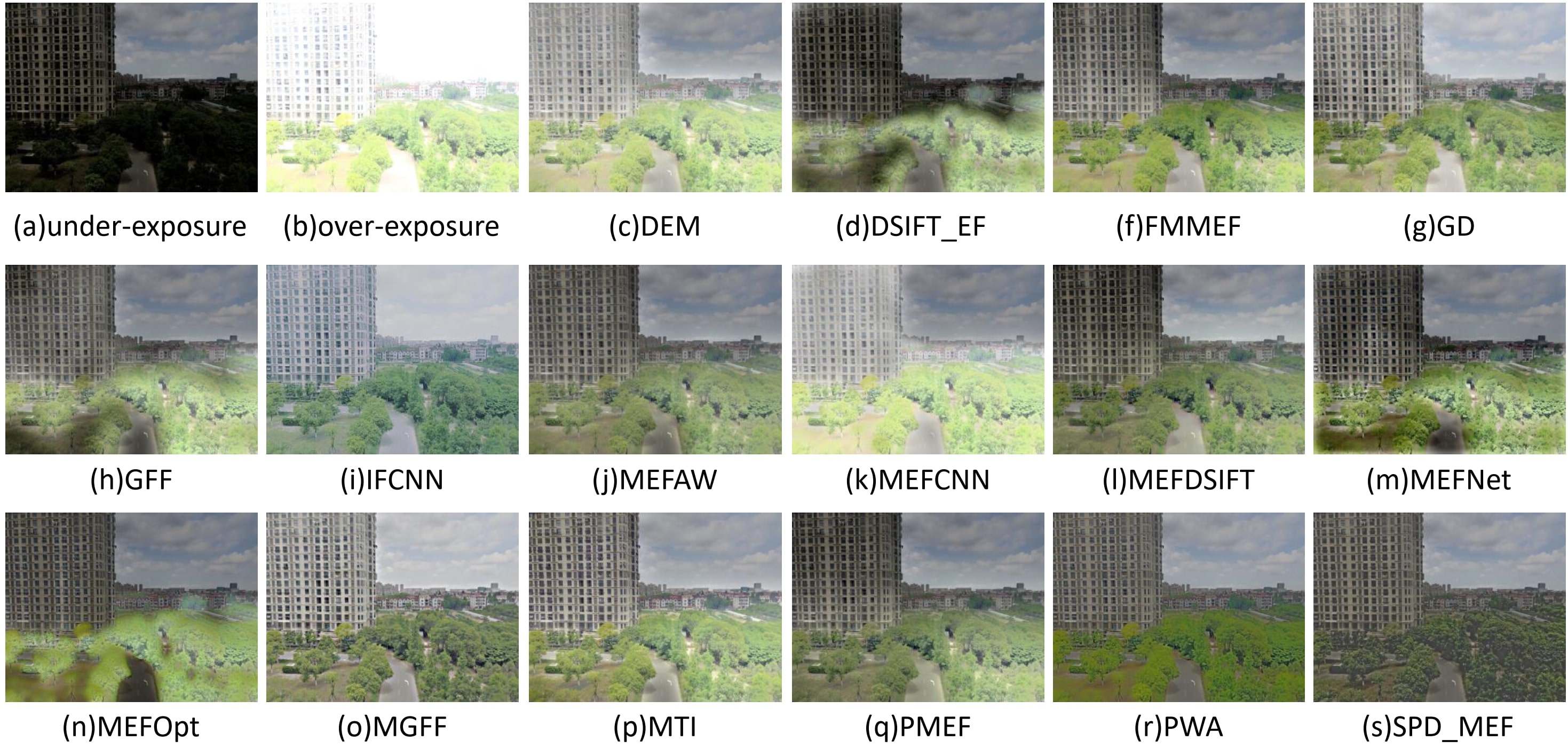}
	\caption{The qualitative performance comparison for the \textit{buildingRoad} image pair.~(a) the under-exposed image. (b) the over-exposed image.~From (c) to (s) are the fused images from 16 MEF approaches, respectively.}
	\label{fig:buildingroad}
\end{figure}

\subsection{Qualitative performance comparison}
\label{subsec:qualitative}
Figure \ref{fig:buildingroad} shows the fused images on the outdoor \textit{buildingRoad} image pair.~It can be seen that several algorithms introduce severe artifacts in the fused images, namely DSIFT\_EF, GFF, MEFAW, MEFNet, MEFOpt.~MTI also introduces small artifact on the road.~Besides, some fused images have strong color distortion, including those produced by IFCNN, PWA and SPD\_MEF.~Furthermore, the fused images given by DEM and MEFCNN are still over-exposed at the bottom half of the image.~Therefore, in this case, FMMEF, GD, MEFDSIFT, MGFF, PMEF give relative better fused images.~Among these give methods, the buildings in the fused images produced by FEMEF, MEFDSIFT and PMEF are relatively darker than those of others and the turn-left sign on the road in the fused image produced by GD is not very clear.~In summary, MGFF gives the best qualitative result on the \textit{buildingRoad} image pair.

\begin{figure}[H]
	\centering
	\includegraphics[width=\textwidth]{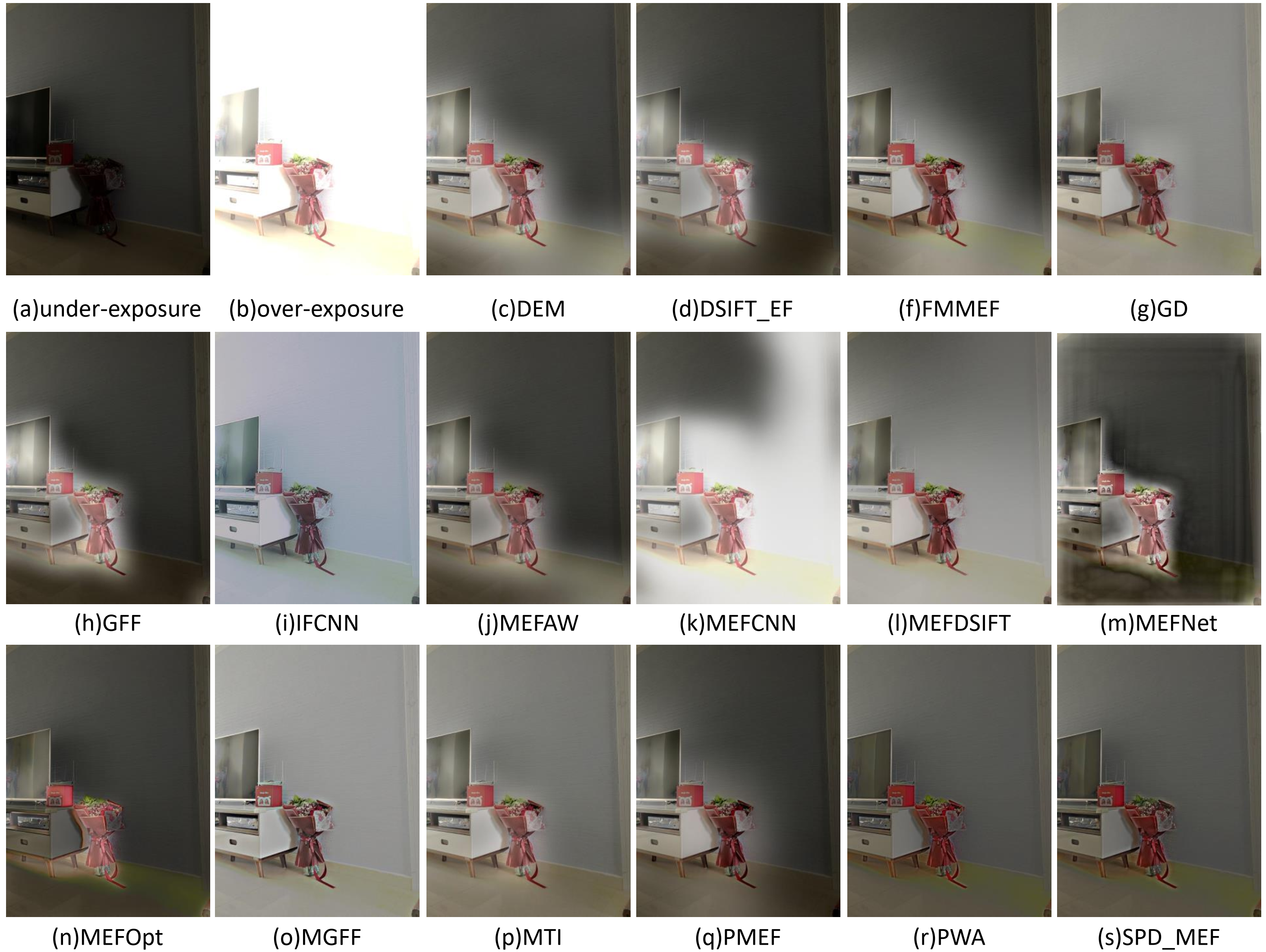}
	\caption{The qualitative performance comparison for the \textit{flower} image pair.~(a) the under-exposed image. (b) the over-exposed image.~From (c) to (s) are the fused images from 16 MEF approaches, respectively.}
	\label{fig:flower}
\end{figure}

\begin{figure}[H]
	\centering
	\includegraphics[width=11cm]{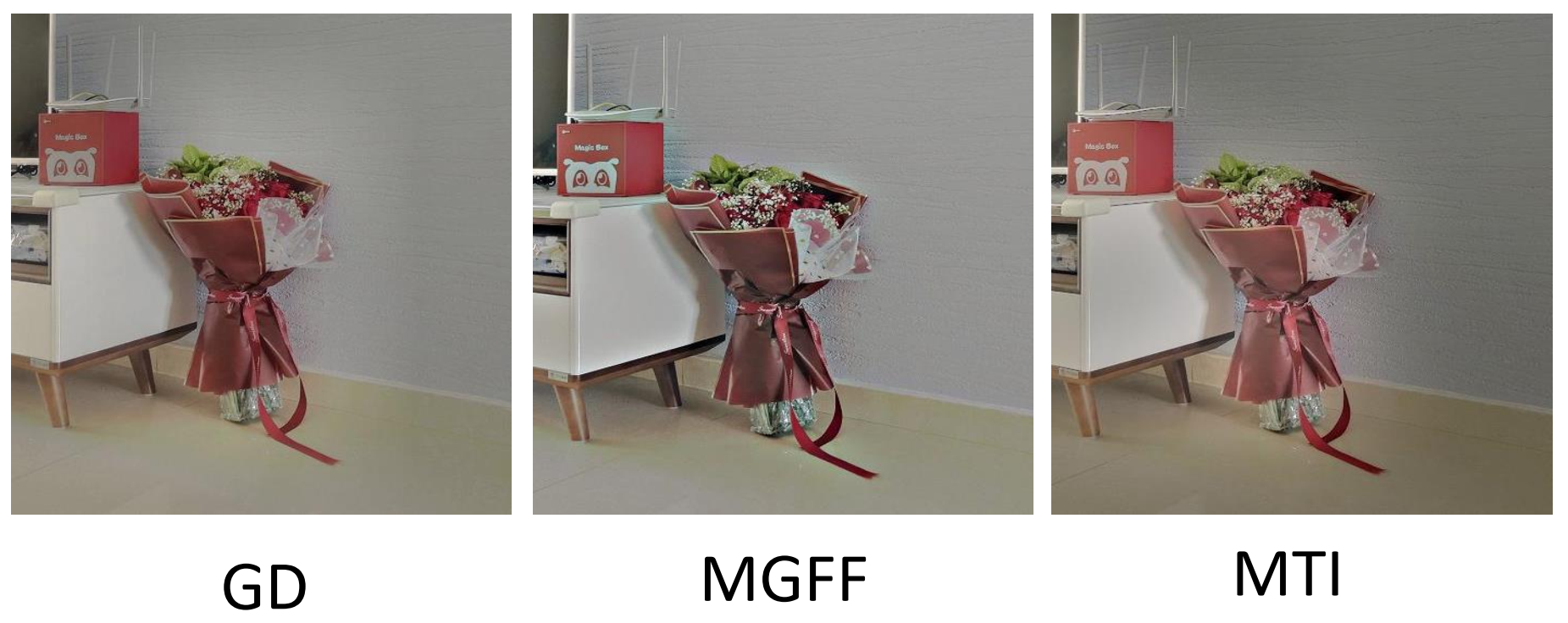}
	\caption{The zoomed-in plots of the \textit{flower} image pair produced by GD, MGFF and MTI, respectively.}
	\label{fig:zoomed-in}
\end{figure}

Figure \ref{fig:flower} illustrates the fused images on an indoor \textit{flower} image pair.~As can be seen, in this case most algorithms produce severe artifacts in the fused images, including DEM, DFSIFT\_EF, FMMEF, GFF, MEFAW, MEFCNN, MEFDSIFT, MEFNet, MEFOpt, PMEF, which means this is a very challenging case.~Besides, IFCNN introduces color distortion in the fused image.~Furthermore,  the fused images obtained by PWA and SPD\_MEF are still a bit under-exposed since the images are not very bright.~Generally speaking, GD, MGFF and MTI perform relative better on this image pair.~Among these three methods, MGFF produces most uniform intensity around the flower as shown in the zoomed-in plot in Fig.~\ref{fig:zoomed-in}.~Besides, the fused image provided by MTI is relative dark.~Therefore, in this \textit{flower} case, MGFF produces the best qualitative result.

\subsection{Quantitative performance comparison}
Table \ref{table:metrics_average} shows the average value of each metric of 20 MEF algorithms.~From the table, it can be seen that the top three algorithms in terms of quantitative results are MGFF, IFCNN and FMMEM, respectively.~Specifically, MGFF obtains the best quantitative performance by having 4 best values, 2 second best values and 2 third best values.~IFCNN shows the second best performance by obtaining 3 best values and 3 second best values.~FMMEM ranked third by achieving 3 best values, 3 second best values and 2 third best values.~Another deep learning-based methods, i.e.~MEFNet shows the same overall quantitative performance with PMFE, following the FMMEF.~However, MEFCNN does not show very competitive performances.~From this table one can see that compared to conventional MEF algorithms, the best deep learning-based algorithms can show competitive performances.~However, some deep learning-based algorithms do not work very well.

	\begin{sidewaystable}
	\centering
	\caption{Average evaluation metric values of all methods on the whole MEFB dataset (100 image pairs).~Note that the metrics are grouped into different categories~The best three values in each metric are denoted in \xz{red}, \xzg{green} and \xzb{blue}, respectively.~The three numbers after the name of each method denote the number of best value, second best value and third best value, respectively.~Best viewed in color.} 
	\scriptsize
	\renewcommand\arraystretch{1.5}
	\begin{tabular}{l|l|lllllll|llll}
		\hline
		&	& \multicolumn{7}{c|}{Information theory-based metrics}  & \multicolumn{4}{c}{Human perception inspired metrics}   \\   \hline     
		\multirow{13}{*}{Conventional}	 &	Method      & CE  & EN  & FMI  & NMI  & PSNR  &  $Q_{NCIE}$ & TE  & $Q_{CB}$ & $Q_{CV}$ & VIF &   \\ \hline
		&	DEM (0,2,1)   & 3.1604 & 7.3129 & \xzg{0.8982} & 0.5463 & 56.8877 & 0.8142 & 60819.05 & 0.4311 & 663.4757 & 0.8011 & \\
		
		&	DSIFT\_EF (2,0,1) & \xzb{2.7380} & 7.3492 & 0.8954 & 0.5405 & 56.6959 & 0.8142 & 105449.4 & 0.4653 & 838.4934 & 0.7273 & \\
		
		&	\xzb{FMMEF (3,2,2)} & 2.9387 & \xzb{7.3714} & 0.8977 & 0.4739 & 56.9473 & 0.8119 & 53253.02 & 0.4516 & 667.487 & \xzg{0.8877} & \\
		
		&	GD (0,0,1) & 3.8027 & 7.2498 & 0.8879 & 0.5506 & 57.1590 & 0.8128 & 28079.4 & 0.4283 & 349.5873 & \xzb{0.8573} &\\
		
		&	GFF (0,1,0) & 3.1698 & \xzg{7.3888} & 0.8869 & 0.4690 & 56.6231 & 0.8121 & 60087.32 & 0.4479 & 942.4572 & 0.7715 &\\
		
		&	MEFOpt (1,2,1) & 3.2736 & 7.1861 & 0.8950 & 0.5860 & 56.8099 & 0.8157 & \xz{141884.8}& \xzb{0.4668} & 711.2798 & 0.7120 &\\
		
		&	MEFAW (0,0,1) & 3.1541 & 7.1993 & 0.8947 & 0.5072 & 56.9615 & 0.8129 & \xzb{109847} & 0.4411 & 698.1415 & 0.7433& \\
		
		&	MEFDSIFT (0,1,0) & \xzg{2.6914} & 7.2536 & 0.8969 & 0.5270& 57.0648 & 0.8128 & 47867.87  & 0.4149 & 531.7647 & 0.7877& \\
		
		&	\xz{MGFF (4,2,2)} & 2.9908 & 7.1198 & 0.8904 & 0.6076 & \xzb{57.2036} & 0.813+ & 9760.439 & \xzg{0.4670} & \xzb{309.5305} & \xz{1.0220}& \\
		
		&	MTI (0,0,2) & 3.3647 & 7.1363 & 0.8918 & 0.5255 & 57.0738 & 0.8129 & 69923.12& 0.4427 & 505.5107 & 0.8089 &\\
		
		&	PMEF (2,1,1) & 2.8755 & 7.3036 & \xz{0.8984} & 0.5325 & 56.9669 & 0.8138 & 70062.18  & 0.4535 & 660.2199 & 0.8368& \\
		
		&	PWA (1,4,2) & 2.9940 & 7.0591 & \xzb{0.8977} & \xzg{0.7530} & \xzg{57.2625} & \xz{0.8198} & 86198.74 & 0.4462 & \xzg{281.2296} & 0.7276& \\
		
		&	SPD\_MEF (1,0,4) & 3.2153 & 7.1234 & 0.8878 & \xzb{0.6954} & 57.1954 & \xzb{0.8178} & 16169.97 & 0.4545 & 350.439 & 0.7824& \\ 	\hline
		
		\multirow{3}{*}{DL-based}	&	\xzg{IFCNN (3,3,0)} & 3.4211 & 7.0339 & 0.8832 & \xz{0.7724} & \xz{57.2855} & \xzg{0.8187} & 64093.55 & 0.4109 & \xz{243.9483} & 0.7032 &\\
		
		&	MEFCNN (1,1,1) & \xz{2.6493} & 7.2566 & 0.8967 & 0.5982 & 56.8182 & 0.8153 & \xzg{114507.3} & 0.4253 & 741.3528 & 0.7308& \\
		
		&	MEFNet (2,1,1) & 3.0453 & \xz{7.3896} & 0.8904 & 0.5978 & 56.6844 & 0.8166 & 77594.44 & \xz{0.4805} & 592.1556 & 0.8486& \\  \hline \hline 
		
		&	&\multicolumn{7}{c|}{Image feature-based metrics} & \multicolumn{4}{c}{Structural similarity-based metrics}   \\   \hline   
		\multirow{13}{*}{Conventional} &	Method      & AG  & EI & $Q^{AB/F}$ & $Q_P$  & SD  & SF & &$Q_C$ & $Q_W$& $Q_Y$ & MEF-SSIM    \\ \hline
		
		&	DEM (0,2,1)  & 5.5224 & 54.8931 & \xzg{0.6982} & 0.6434 & 53.3167 & 18.8680& & 0.6583 & 0.8834 & \xzb{0.7502} & 0.9593  \\
		
		&	DSIFT\_EF (2,0,1) & 5.0424 & 50.3853 & 0.6782 & 0.6411 & 51.357 & 17.1133& & \xz{0.6695} & 0.8619 & \xz{0.7577} & 0.9483 \\
		
		&	\xzb{FMMEF (3,2,2)} & 5.5832 & 55.8741 & \xz{0.7049} & \xz{0.6546} & \xzb{56.6406} & 19.0416 && 0.6414 & \xz{0.9072} & 0.7213& \xzg{0.9741}  \\
		
		&	GD (0,0,1) &  5.5181 & 54.4657 & 0.6720 & 0.6133 & 52.4781 & 17.9503& & 0.5915 & 0.8494 & 0.6523 & 0.9647  \\
		
		&	GFF (0,1,0) & 5.5591 & 55.5594& 0.6178 & 0.5246 & 52.7002 & 18.6044& & 0.5995 & 0.8771 & 0.6560 & 0.9431 \\
		
		&	MEFOpt (1,2,1) & 5.7079 & 56.8206 & 0.6880 & 0.6070 & 50.2844 & 19.2862& & \xzg{0.6638} & 0.8871 & \xzg{0.7508} & 0.9368 \\
		
		&	MEFAW (0,0,1) &  4.9771 & 49.7620 & 0.6887 & 0.6459 & 49.0940 & 16.7862& & 0.6488 & 0.8706 & 0.7247 & 0.9628 \\
		
		&	MEFDSIFT (0,1,0) & 5.1625 & 51.5098 & 0.6615 & 0.6311 & 54.5330 & 17.8953& & 0.6271 & 0.8195 & 0.7158 & 0.9612 \\
		
		&	\xz{MGFF (4,2,2)} & \xz{6.4808} & \xz{64.8304} & 0.6402 & 0.6337 & \xzg{58.3029} & \xz{22.4312} && 0.5596 & 0.8465 & 0.6105 & 0.9553 \\
		
		&	MTI (0,0,2) &5.5736 & 55.2379 & \xzb{0.6974} & \xzb{0.6468} & 47.8058 & 18.6622 && 0.6425 & 0.8780 & 0.7205 & 0.9664 \\
		
		&	PMEF (2,1,1) &  5.4662 & 54.6308 & 0.6955 & \xzg{0.6501} & 55.6796 & 18.7435 && 0.6547 & \xzb{0.8904} & 0.7447 & \xz{0.9745} \\
		
		&	PWA (1,4,2) & 5.5049 & 54.8211 & 0.6961 & 0.6393& 56.2055 & 18.8545& & 0.6445 & \xzg{0.8959} & 0.7281 & \xzb{0.9664}  \\
		
		&	SPD\_MEF (1,0,4) &  5.9146 & \xzb{58.5438} & 0.6321 & 0.6153 & \xz{58.5788} & \xzb{20.7757}& & 0.6181 & 0.8034 & 0.6845 & 0.9382  \\ 	\hline
		
		\multirow{3}{*}{DL-based}	&	\xzg{IFCNN (3,3,0)} &  \xzg{5.9644} & 58.1424 & 0.5965 & 0.5586 & 56.0002 & \xzg{21.0352}& & 0.5526 & 0.8354 & 0.6085 & 0.9438  \\
		
		&	MEFCNN (1,1,1) & 4.9462& 49.4188 & 0.6715 & 0.6387 & 52.3796 & 16.8925& & \xzb{0.6587} & 0.8387& 0.7467 & 0.9374 \\
		
		&	MEFNet (2,1,1) &  \xzb{5.9428} & \xzg{59.5320} & 0.6743 & 0.59247 & 55.7468 & 20.1128 && 0.6549 & 0.8657& 0.7334 & 0.9148 \\
		
		\hline
	\end{tabular}%
	\label{table:metrics_average}%
	\end{sidewaystable}

One interesting thing can be found from Table \ref{table:metrics_average} is that, the top three MEF algorithms show very different performances in different kinds of metrics.~To be more specific, MGFF mainly performs well in image feature-based metrics and human perception inspired metrics, while IFCNN mainly shows good performances in information theory-based metrics and in some of the image feature-based metrics (AG, SF) and one human perception inspired metric ($Q_{CV}$).~In contrast, FMMEF mainly exhibits good performances in image feature-based metrics and structural similarity-based ones.~This phenomenon 
clearly indicates that different kinds of metrics can examine the MEF approaches from different aspects, therefore it is essential to evaluate MEF algorithms using metrics covering more kinds.~It is not a good idea to evaluate MEF algorithms using very few metrics which only cover a part of kinds as shown in Table \ref{table:published-some}.

From Section \ref{subsec:qualitative} and this Section one can see that the qualitative results and quantitative results are consistent to some extent.~For instance, MGFF obtains the best performance in terms of both qualitative and quantitative results.~In particular, MGFF performs well in human perception inspired metrics.~However, it should be mentioned that the quantitative performances and qualitative performances are not always consistent.~For example, the deep learning-based algorithm IFCNN obtains the second best overall quantitative results, but its fused images have severe color distortion.~Therefore, it is necessary to use both quantitative and qualitative methods when evaluating MEF algorithms.~Another thing should be mentioned is that deep learning-based methods do not provide good qualitative results according our experimental results.

\subsection{Running time comparison}
\label{subsec:runtime}
To compare the computational costs of MEF algorithms, the running time of the algorithms integrated in MEFB are listed in Table \ref{subsec:runtime}.~As can be seen, the running time varies significantly from one method to another.~Generally speaking, deep learning-based MEF methods are efficient with the help of GPU.~Besides, patch-based methods are more expensive than their pixel-based counterparts.~The optimization-based method MEFOpt takes around seven minutes to fuse one image pair, which is too slow.~This is due to the high computational cost induced by iterative optimization in the this method.~Regarding pixel-based, patch-based, edge-preserving-based and multi-scale-based methods, it is hard to say which kind is more efficient because the running time is highly dependent on the implementations.~But their running time are generally in the same order of magnitude.

\begin{table}
	\begin{center}
		\caption{Average running time of algorithms in MEFB (seconds per image pair)}
		\label{table:runtime}
		\scriptsize
		\begin{tabular}{llllll}		
			\hline
			Method      & Running time     & Category & Method      & Running time   & Category    \\ \hline
			DSIFT\_EF \cite{liu2015dense} & 1.26 & Spatial-domain (pixel-based) &   MTI \cite{yang2018multi} & 2.78 &  Multi-scale-based	\\	
			MEFAW \cite{lee2018multi}	  &0.57 &  Spatial-domain (pixel-based) & GD \cite{paul2016multi}&1.76 & Multi-scale-based     \\ 
			MEFDSIFT \cite{hayat2019ghost} & 1.53&Spatial-domain (pixel-based)      & DEM \cite{wang2019detail}   & 0.71 &  	Multi-scale-based  \\
			PMEF \cite{liu2019perceptual} & 0.64 &  Spatial-domain (pixel-based)  &FMMEF \cite{li2020fast}	& 0.73  & Multi-scale-based	 		 \\
			SPD-MEF \cite{ma2017robust}& 3.48 & Spatial-domain (patch-based)  & MEFOpt \cite{ma2017multib}	&419.27 & Optimization-based  	       \\
			PWA \cite{ma2015multi}  & 2.61 &  Spatial-domain (patch-based)	 & MEFNet \cite{ma2020deep}	  & 0.42 &DL-based\\	
			MGFF \cite{bavirisetti2019multi}   	&2.86 &	Edge-preserving-based &MEFCNN \cite{li2018multib} & 0.71 &  DL-based  \\
			GFF \cite{li2013image}& 0.88 & Edge-preserving-based  &	IFCNN \cite{zhang2020ifcnn}	& 0.67 &DL-based\\ \hline
		\end{tabular}
	\end{center}
\end{table}

\section{Discussions}
\label{sec:discussions}

\subsection{About the test set in MEFB}
MEFB is proposed to fill the gap of lacking platforms for fair performance evaluation in MEF.~The proposed MEFB to MEF, is like the OTB and VOT to object tracking.~As indicated by Table 1, previously
different images and metrics were utilized in MEF literature, making it difficult to know the real performances of
algorithms and the state-of-the-arts.~What we really care in this work is solving this issue.~Therefore, the dataset in MEFB is a test set.~As listed in Table 1, 100 image pairs are much larger than those utilized in MEF literatures.~However, more image pairs would be better, therefore in future we will add more image pairs into MEFB.

\subsection{About the performances of deep learning-based MEF methods}
From the experimental results, one can see that although the IFCNN shows competitive results compared to conventional approaches, the other two examined deep learning-based methods fail to beat many conventional algorithms.~However, these algorithms were all claimed to have state-of-the-art performances.~The possible reason of this is the lack of benchmark previously, thus it is difficult to fully test the performance of MEF algorithms from different point of view by using various kinds of evaluation metrics.~The proposed MEFB can help to solve this issue in the future by providing a test bed which is easily to use.

\section{Conclusions}
\label{sec:conclusion}
In this paper, we proposed the MEFB, which is a multi-exposure image fusion benchmark aiming at providing a platform as well as a toolkit for comprehensive performance comparison of MEF algorithms.~Currently, MEFB consists of 100 multi-exposure image pairs, 16 MEF algorithms, 20 evaluation metrics, 1600 fused images and a software toolkit.~To the best of our knowledge, this is the first benchmark in the field of multi-exposure image fusion.~MEFB can be used conveniently and it can be easily extended to include more images, methods and evaluation metrics. 

We also conducted extensive experiments using MEFB to compare the performances of MEF methods in order to understand the state-of-the-art of this field.~Based on the results, we find that the deep learning-based method are not dominant in the field of MEF since some deep learning-based algorithms do not show competitive performances compared to conventional algorithms.~Also, we find that it is essential to evaluate MEF algorithms using different kinds of metrics as the performances of MEF approaches vary significantly from one metric to another.~This also indicates that when designing loss functions for deep learning-based MEF algorithms, several kinds of metrics could be considered together in the loss function, instead of using MEF-SSIM only as most deep learning-based MEF method did.

We expect that the proposed MEFB will make it feasible for fair and comprehensive performance comparison of MEF approaches and 
can serve as a good tool for for researchers in this field.

\section*{Acknowledgment}
The author would like to the corresponding authors of all integrated image fusion algorithms for providing the source codes.

\section*{References}

\biboptions{numbers,sort&compress}
\bibliography{../../../../xingchen}

\begin{thebibliography}{10}
\expandafter\ifx\csname url\endcsname\relax
  \def\url#1{\texttt{#1}}\fi
\expandafter\ifx\csname urlprefix\endcsname\relax\def\urlprefix{URL }\fi
\expandafter\ifx\csname href\endcsname\relax
  \def\href#1#2{#2} \def\path#1{#1}\fi

\bibitem{shen2011generalized}
R.~Shen, I.~Cheng, J.~Shi, A.~Basu, Generalized random walks for fusion of
  multi-exposure images, IEEE Transactions on Image Processing 20~(12) (2011)
  3634--3646.

\bibitem{yang2018multi}
Y.~Yang, W.~Cao, S.~Wu, Z.~Li, Multi-scale fusion of two large-exposure-ratio
  images, IEEE Signal Processing Letters 25~(12) (2018) 1885--1889.

\bibitem{ma2015multi}
K.~Ma, Z.~Wang, Multi-exposure image fusion: A patch-wise approach, in: 2015
  IEEE International Conference on Image Processing (ICIP), IEEE, 2015, pp.
  1717--1721.

\bibitem{li2020fast}
H.~Li, K.~Ma, H.~Yong, L.~Zhang, Fast multi-scale structural patch
  decomposition for multi-exposure image fusion, IEEE Transactions on Image
  Processing 29 (2020) 5805--5816.

\bibitem{lee2018multi}
S.-h. Lee, J.~S. Park, N.~I. Cho, A multi-exposure image fusion based on the
  adaptive weights reflecting the relative pixel intensity and global gradient,
  in: 2018 25th IEEE International Conference on Image Processing (ICIP), IEEE,
  2018, pp. 1737--1741.

\bibitem{ma2017multib}
K.~Ma, Z.~Duanmu, H.~Yeganeh, Z.~Wang, Multi-exposure image fusion by
  optimizing a structural similarity index, IEEE Transactions on Computational
  Imaging 4~(1) (2017) 60--72.

\bibitem{lahoud2019fast}
F.~Lahoud, S.~S{\"{u}}sstrunk, {Fast and Efficient Zero-Learning Image Fusion}
  (2019) 1--13\href {http://arxiv.org/abs/1905.03590}
  {\path{arXiv:1905.03590}}.

\bibitem{prabhakar2017deepfuse}
K.~R. Prabhakar, V.~S. Srikar, R.~V. Babu, Deepfuse: A deep unsupervised
  approach for exposure fusion with extreme exposure image pairs, in: 2017 IEEE
  International Conference on Computer Vision (ICCV). IEEE, 2017, pp.
  4724--4732.

\bibitem{deng2020deep}
X.~Deng, P.~L. Dragotti, Deep convolutional neural network for multi-modal
  image restoration and fusion, IEEE Transactions on Pattern Analysis and
  Machine Intelligence.

\bibitem{li2013image}
S.~Li, X.~Kang, J.~Hu, Image fusion with guided filtering, IEEE Transactions on
  Image processing 22~(7) (2013) 2864--2875.

\bibitem{ma2017robust}
K.~Ma, H.~Li, H.~Yong, Z.~Wang, D.~Meng, L.~Zhang, Robust multi-exposure image
  fusion: A structural patch decomposition approach, IEEE Transactions on Image
  Processing 26~(5) (2017) 2519--2532.

\bibitem{jung2020unsupervised}
H.~Jung, Y.~Kim, H.~Jang, N.~Ha, K.~Sohn, {Unsupervised Deep Image Fusion with
  Structure Tensor Representations}, IEEE Transactions on Image Processing 29
  (2020) 3845--3858.
\newblock \href {http://dx.doi.org/10.1109/TIP.2020.2966075}
  {\path{doi:10.1109/TIP.2020.2966075}}.

\bibitem{zhang2020ifcnn}
Y.~Zhang, Y.~Liu, P.~Sun, H.~Yan, X.~Zhao, L.~Zhang, {IFCNN: A general image
  fusion framework based on convolutional neural network}, Information Fusion
  54~(August 2018) (2020) 99--118.

\bibitem{xu2020fusiondn}
H.~Xu, J.~Ma, Z.~Le, J.~Jiang, X.~Guo, {Fusiondn: A unified densely connected
  network for image fusion}, in: Thirty-Fourth AAAI Conference on Artificial
  Intelligence, 2020.

\bibitem{zhang2020PMGI}
H.~Zhang, H.~Xu, Y.~Xiao, X.~Guo, J.~Ma, Rethinking the image fusion: A fast
  unified image fusion network based on proportional maintenance of gradient
  and intensity, in: Proceedings of the AAAI Conference on Artificial
  Intelligence, 2020.

\bibitem{wu2013online}
Y.~Wu, J.~Lim, M.-H. Yang, Online object tracking: A benchmark, in: Proceedings
  of the IEEE conference on computer vision and pattern recognition, 2013, pp.
  2411--2418.

\bibitem{wu2015object}
Y.~Wu, J.~Lim, M.-H. Yang, Object tracking benchmark, IEEE Transactions on
  Pattern Analysis and Machine Intelligence 37~(9) (2015) 1834--1848.

\bibitem{li2019rgb}
C.~Li, X.~Liang, Y.~Lu, N.~Zhao, J.~Tang, Rgb-t object tracking: benchmark and
  baseline, Pattern Recognition (2019) 106977.

\bibitem{zhang2020dsiammft}
X.~Zhang, P.~Ye, S.~Peng, J.~Liu, G.~Xiao, {DSiamMFT: An RGB-T fusion tracking
  method via dynamic Siamese networks using multi-layer feature fusion}, Signal
  Processing: Image Communication (2020) 115756.

\bibitem{zhang2020object}
X.~Zhang, P.~Ye, H.~Leung, K.~Gong, G.~Xiao, Object fusion tracking based on
  visible and infrared images: A comprehensive review, Information Fusion 63
  (2020) 166--187.

\bibitem{james2014medical}
A.~P. James, B.~V. Dasarathy, Medical image fusion: A survey of the state of
  the art, Information Fusion 19 (2014) 4--19.

\bibitem{liu2017a}
Y.~Liu, X.~Chen, J.~Cheng, H.~P. I.~F. (Fusion), U.~2017, {A medical image
  fusion method based on convolutional neural networks}, Information Fusion
  (Fusion), 2017 20th International Conference on (2017) 1--7.

\bibitem{liu2017multi}
Y.~Liu, X.~Chen, H.~Peng, Z.~Wang, Multi-focus image fusion with a deep
  convolutional neural network, Information Fusion 36 (2017) 191--207.

\bibitem{zhang2020multifocus}
X.~Zhang, Multi-focus image fusion: A benchmark (2020).
\newblock \href {http://arxiv.org/abs/2005.01116} {\path{arXiv:2005.01116}}.

\bibitem{ghassemian2016review}
H.~Ghassemian, A review of remote sensing image fusion methods, Information
  Fusion 32 (2016) 75--89.

\bibitem{ye2018fusioncnn}
F.~Ye, X.~Li, X.~Zhang, {FusionCNN : a remote sensing image fusion algorithm
  based on deep convolutional neural networks}.

\bibitem{ma2015perceptual}
K.~Ma, K.~Zeng, Z.~Wang, Perceptual quality assessment for multi-exposure image
  fusion, IEEE Transactions on Image Processing 24~(11) (2015) 3345--3356.

\bibitem{ma2016infrared}
J.~Ma, C.~Chen, C.~Li, J.~Huang, Infrared and visible image fusion via gradient
  transfer and total variation minimization, Information Fusion 31 (2016)
  100--109.

\bibitem{zhang2020vifb}
X.~Zhang, P.~Ye, G.~Xiao, {VIFB: A Visible and Infrared Image Fusion
  Benchmark}, in: Proceedings of the IEEE/CVF Conference on Computer Vision and
  Pattern Recognition Workshops, 2020.

\bibitem{yin2018tensor}
H.~Yin, Tensor sparse representation for 3-d medical image fusion using
  weighted average rule, IEEE Transactions on Biomedical Engineering 65~(11)
  (2018) 2622--2633.

\bibitem{hill2016perceptual}
P.~Hill, M.~E. Al-Mualla, D.~Bull, Perceptual image fusion using wavelets, IEEE
  transactions on image processing 26~(3) (2016) 1076--1088.

\bibitem{he2010multimodal}
C.~He, Q.~Liu, H.~Li, H.~Wang, Multimodal medical image fusion based on ihs and
  pca, Procedia Engineering 7 (2010) 280--285.

\bibitem{liu2015general}
Y.~Liu, S.~Liu, Z.~Wang, A general framework for image fusion based on
  multi-scale transform and sparse representation, Information Fusion 24 (2015)
  147--164.

\bibitem{wan2011application}
T.~Wan, Z.~Qin, An application of compressive sensing for image fusion,
  International Journal of Computer Mathematics 88~(18) (2011) 3915--3930.

\bibitem{jin2017survey}
X.~Jin, Q.~Jiang, S.~Yao, D.~Zhou, R.~Nie, J.~Hai, K.~He, A survey of infrared
  and visual image fusion methods, Infrared Physics \& Technology 85 (2017)
  478--501.

\bibitem{li2017pixel}
S.~Li, X.~Kang, L.~Fang, J.~Hu, H.~Yin, Pixel-level image fusion: A survey of
  the state of the art, Information Fusion 33 (2017) 100--112.

\bibitem{liu2018deep}
Y.~Liu, X.~Chen, Z.~Wang, Z.~J. Wang, R.~K. Ward, X.~Wang, Deep learning for
  pixel-level image fusion: Recent advances and future prospects, Information
  Fusion 42 (2018) 158--173.

\bibitem{ma2019infrared}
J.~Ma, Y.~Ma, C.~Li, Infrared and visible image fusion methods and
  applications: A survey, Information Fusion 45 (2019) 153--178.

\bibitem{hermessi2018convolutional}
H.~Hermessi, O.~Mourali, E.~Zagrouba, Convolutional neural network-based
  multimodal image fusion via similarity learning in the shearlet domain,
  Neural Computing and Applications (2018) 1--17.

\bibitem{yan2018unsupervised}
X.~Yan, S.~Z. Gilani, H.~Qin, A.~Mian, Unsupervised deep multi-focus image
  fusion, arXiv preprint arXiv:1806.07272.

\bibitem{xia2018novel}
K.~Xia, H.~Yin, J.~Wang, A novel improved deep convolutional neural network
  model for medical image fusion, Cluster Computing (2018) 1--13.

\bibitem{ma2019fusiongan}
J.~Ma, W.~Yu, P.~Liang, C.~Li, J.~Jiang, {FusionGAN: A generative adversarial
  network for infrared and visible image fusion}, Information Fusion 48~(June
  2018) (2019) 11--26.

\bibitem{liu2018infrared}
Y.~Liu, X.~Chen, J.~Cheng, H.~Peng, Z.~Wang, Infrared and visible image fusion
  with convolutional neural networks, International Journal of Wavelets,
  Multiresolution and Information Processing 16~(03) (2018) 1850018.

\bibitem{li2019densefuse}
H.~Li, X.~Wu, Densefuse: A fusion approach to infrared and visible images, IEEE
  Transactions on Image Processing 28~(5) (2019) 2614--2623.

\bibitem{fang2019perceptual}
Y.~Fang, H.~Zhu, K.~Ma, Z.~Wang, S.~Li, Perceptual evaluation for
  multi-exposure image fusion of dynamic scenes, IEEE Transactions on Image
  Processing 29 (2019) 1127--1138.

\bibitem{cai2018learning}
J.~Cai, S.~Gu, L.~Zhang, Learning a deep single image contrast enhancer from
  multi-exposure images, IEEE Transactions on Image Processing 27~(4) (2018)
  2049--2062.

\bibitem{liu2019perceptual}
X.~Liu, Y.~Liu, C.~Zhu, Perceptual multi-exposure image fusion, IEEE
  Transactions on Multimedia.

\bibitem{zeng2014perceptual}
K.~Zeng, K.~Ma, R.~Hassen, Z.~Wang, Perceptual evaluation of multi-exposure
  image fusion algorithms, in: 2014 Sixth International Workshop on Quality of
  Multimedia Experience (QoMEX), IEEE, 2014, pp. 7--12.

\bibitem{wang2019detail}
Q.~Wang, W.~Chen, X.~Wu, Z.~Li, Detail-enhanced multi-scale exposure fusion in
  yuv color space, IEEE Transactions on Circuits and Systems for Video
  Technology.

\bibitem{liu2015dense}
Y.~Liu, Z.~Wang, Dense sift for ghost-free multi-exposure fusion, Journal of
  Visual Communication and Image Representation 31 (2015) 208--224.

\bibitem{paul2016multi}
S.~Paul, I.~S. Sevcenco, P.~Agathoklis, Multi-exposure and multi-focus image
  fusion in gradient domain, Journal of Circuits, Systems and Computers 25~(10)
  (2016) 1650123.

\bibitem{li2018multib}
H.~Li, L.~Zhang, {MULTI-EXPOSURE FUSION WITH CNN FEATURES}, 2018 25th IEEE
  International Conference on Image Processing (ICIP) (2018) 1723--1727.

\bibitem{hayat2019ghost}
N.~Hayat, M.~Imran, Ghost-free multi exposure image fusion technique using
  dense sift descriptor and guided filter, Journal of Visual Communication and
  Image Representation 62 (2019) 295--308.

\bibitem{ma2020deep}
K.~Ma, Z.~Duanmu, H.~Zhu, Y.~Fang, Z.~Wang, Deep guided learning for fast
  multi-exposure image fusion, IEEE Transactions on Image Processing.

\bibitem{bavirisetti2019multi}
D.~P. Bavirisetti, G.~Xiao, J.~Zhao, R.~Dhuli, G.~Liu, Multi-scale guided image
  and video fusion: A fast and efficient approach, Circuits, Systems, and
  Signal Processing 38~(12) (2019) 5576--5605.

\bibitem{li2018multi}
H.~Li, X.-J. Wu, Multi-focus noisy image fusion using low-rank representation,
  arXiv preprint arXiv:1804.09325.

\bibitem{liu2012objective}
Z.~Liu, E.~Blasch, Z.~Xue, J.~Zhao, R.~Laganiere, W.~Wu, Objective assessment
  of multiresolution image fusion algorithms for context enhancement in night
  vision: A comparative study, IEEE Transactions on Pattern Analysis and
  Machine Intelligence 34 (2012) 94--109.
\newblock \href
  {http://dx.doi.org/http://doi.ieeecomputersociety.org/10.1109/TPAMI.2011.109}
  {\path{doi:http://doi.ieeecomputersociety.org/10.1109/TPAMI.2011.109}}.

\bibitem{bulanon2009image}
D.~M. Bulanon, T.~Burks, V.~Alchanatis, Image fusion of visible and thermal
  images for fruit detection, Biosystems Engineering 103~(1) (2009) 12--22.

\bibitem{aardt2008assessment}
V.~Aardt, Jan, Assessment of image fusion procedures using entropy, image
  quality, and multispectral classification, Journal of Applied Remote Sensing
  2~(1) (2008) 023522.

\bibitem{hossny2008comments}
M.~Hossny, S.~Nahavandi, D.~Creighton, Comments on'information measure for
  performance of image fusion', Electronics letters 44~(18) (2008) 1066--1067.

\bibitem{jagalingam2015review}
P.~Jagalingam, A.~V. Hegde, A review of quality metrics for fused image,
  Aquatic Procedia 4~(Icwrcoe) (2015) 133--142.

\bibitem{wang2005nonlinear}
Q.~Wang, Y.~Shen, J.~Q. Zhang, A nonlinear correlation measure for
  multivariable data set, Physica D: Nonlinear Phenomena 200~(3-4) (2005)
  287--295.

\bibitem{wang2008performance}
Q.~Wang, Y.~Shen, J.~Jin, Performance evaluation of image fusion techniques,
  Image fusion: algorithms and applications 19 (2008) 469--492.

\bibitem{cvejic2006image}
N.~Cvejic, C.~Canagarajah, D.~Bull, Image fusion metric based on mutual
  information and tsallis entropy, Electronics letters 42~(11) (2006) 626--627.

\bibitem{cui2015detail}
G.~Cui, H.~Feng, Z.~Xu, Q.~Li, Y.~Chen, Detail preserved fusion of visible and
  infrared images using regional saliency extraction and multi-scale image
  decomposition, Optics Communications 341 (2015) 199 -- 209.

\bibitem{rajalingam2018hybrid}
B.~Rajalingam, R.~Priya, Hybrid multimodality medical image fusion technique
  for feature enhancement in medical diagnosis, International Journal of
  Engineering Science Invention.

\bibitem{xydeas2000objective}
C.~S. Xydeas, P.~V. V., Objective image fusion performance measure, Military
  Technical Courier 36~(4) (2000) 308--309.

\bibitem{zhao2007performance}
J.~Zhao, R.~Laganiere, Z.~Liu, Performance assessment of combinative
  pixel-level image fusion based on an absolute feature measurement,
  International Journal of Innovative Computing, Information and Control 3~(6)
  (2007) 1433--1447.

\bibitem{rao1997fibre}
Y.-J. Rao, In-fibre bragg grating sensors, Measurement science and technology
  8~(4) (1997) 355.

\bibitem{eskicioglu1995image}
A.~M. Eskicioglu, P.~S. Fisher, Image quality measures and their performance,
  IEEE Transactions on communications 43~(12) (1995) 2959--2965.

\bibitem{cvejic2005similarity}
N.~Cvejic, A.~Loza, D.~Bull, N.~Canagarajah, A similarity metric for assessment
  of image fusion algorithms, International journal of signal processing 2~(3)
  (2005) 178--182.

\bibitem{piella2003new}
G.~Piella, H.~Heijmans, A new quality metric for image fusion, in: Proceedings
  2003 International Conference on Image Processing (Cat. No. 03CH37429),
  Vol.~3, IEEE, 2003, pp. III--173.

\bibitem{yang2008novel}
C.~Yang, J.-Q. Zhang, X.-R. Wang, X.~Liu, A novel similarity based quality
  metric for image fusion, Information Fusion 9~(2) (2008) 156--160.

\bibitem{chen2009new}
Y.~Chen, R.~S. Blum, A new automated quality assessment algorithm for image
  fusion, Image and vision computing 27~(10) (2009) 1421--1432.

\bibitem{chen2007human}
H.~Chen, P.~K. Varshney, A human perception inspired quality metric for image
  fusion based on regional information, Information fusion 8~(2) (2007)
  193--207.

\bibitem{han2013new}
Y.~Han, Y.~Cai, Y.~Cao, X.~Xu, {A new image fusion performance metric based on
  visual information fidelity}, Information Fusion 14~(2) (2013) 127--135.
\newblock \href {http://dx.doi.org/10.1016/j.inffus.2011.08.002}
  {\path{doi:10.1016/j.inffus.2011.08.002}}.

\bibitem{qu2002information}
G.~Qu, D.~Zhang, P.~Yan, Information measure for performance of image fusion,
  Electronics letters 38~(7) (2002) 313--315.

\bibitem{haghighat2011non}
M.~B.~A. Haghighat, A.~Aghagolzadeh, H.~Seyedarabi, A non-reference image
  fusion metric based on mutual information of image features, Computers \&
  Electrical Engineering 37~(5) (2011) 744--756.

\bibitem{wang2004image}
Z.~Wang, A.~C. Bovik, H.~R. Sheikh, E.~P. Simoncelli, et~al., Image quality
  assessment: from error visibility to structural similarity, IEEE transactions
  on image processing 13~(4) (2004) 600--612.

\end{thebibliography}
\end{document}